\documentclass[letterpaper]{article} 
\usepackage[preprint]{aaai2027} 
\usepackage[hyphens]{url}  
\usepackage{graphicx} 
\usepackage{natbib}  
\usepackage{caption} 
\usepackage{algorithm}
\usepackage{algorithmic}
\usepackage{multirow}
\usepackage{newfloat}
\usepackage{listings}
\DeclareCaptionStyle{ruled}{labelfont=normalfont,labelsep=colon,strut=off} 
\floatstyle{ruled}
\newfloat{listing}{tb}{lst}{}
\floatname{listing}{Listing}

\usepackage{booktabs}
\usepackage{amsmath}
\usepackage{amssymb}
\newcommand{\model}{\textsc{SPEM}}
\title{SPEM: Structure-Aware Procedural Memory for Efficient Skill Use in LLM Agents}
\author{
    Written by AAAI Press Staff\textsuperscript{\rm 1}\thanks{With help from the AAAI Publications Committee.}\\
    AAAI Style Contributions by Peter Patel Schneider,
    Sunil Issar,\\
    J. Scott Penberthy,
    George Ferguson,
    Hans Guesgen,
    Francisco Cruz\equalcontrib\corresponding,
    Marc Pujol-Gonzalez\equalcontrib\corresponding
}
\author{
    Ran Zhou,
    Zehui Kong,
    Fanyu Lin,
    Jiaheng Liu,
    Zhen Zhang,
    Yuxiang Ren\corresponding
}
\affiliations{
    School of Intelligence Science and Technology, Nanjing University\\
    Correspondence: renyuxiang@nju.edu.cn
}

\begin{document}

\maketitle

\begin{abstract}
LLM-based agents increasingly rely on reusable skills. However, scaling skill libraries makes efficient and accurate skill use difficult. Full skill loading incurs high token and runtime overhead. In contrast, flat semantic retrieval treats skills as independent text entries, ignoring dependencies, execution order, state transitions, and data-flow constraints. We propose \model{}, a lightweight structure-aware procedural memory framework for efficient skill use in LLM-based agents. \model{} maintains Atomic Skill Memory for individual skills and Skill-DAG Memory for reusable compositions derived from successful executions. For a new task, \model{} reuses high-confidence Skill-DAG templates when available. Otherwise, it retrieves atomic skills and invokes Skill-GNN for state- and structure-aware skill selection. Successful executions are written back as new Skill-DAGs, allowing procedural memory to accumulate over time. Experiments on ALFWorld and SkillsBench show that \model{} achieves the best overall effectiveness--efficiency trade-off, reaching 87.1\% ALFWorld success and 38.38\% SkillsBench reward while reducing token and runtime costs against strong baselines.
\end{abstract}


\section{Introduction}

Large language model (LLM) agents are increasingly used as executable systems. They call tools, reuse skills, and interact with complex environments. In software engineering, data analysis, web automation, and embodied tasks, agents rely on reusable skills, such as file parsing, preprocessing, and API calling~\cite{schick2023toolformer,mialon2023augmented,yao2023react}. As skill libraries scale to hundreds or thousands of candidates, the central challenge changes. It shifts from whether useful skills exist to how an agent can retrieve, organize, and reuse the required skills efficiently~\cite{patil2023gorilla,qin2024toolllm,shi2025toolsavvy}.

Existing approaches face four limitations under large-scale skill libraries. First, loading all skill descriptions is infeasible due to high token cost,
context bloat, and runtime overhead, with important details often lost in long
contexts~\cite{liu2024lost}. Second, flat retrieval optimizes semantic relevance
but ignores structural dependencies like prerequisites, order, state
transitions, and data flow across skills~\cite{lewis2020rag,karpukhin2020dpr}.
Third, successful compositions are not stored as reusable procedural memory. Even
when dependency-aware plans are built, they are discarded after execution. This
forces agents to re-derive similar workflows~\cite{yao2023react,shinn2023reflexion}.
Finally, as agent systems are deployed more frequently and model capabilities continue to improve, efficiency becomes a primary concern. Repeated model invocations for retrieval, planning, and reorganization incur unnecessary token consumption, latency, and computational overhead. These limitations call for a framework that can retrieve skills compactly, organize them structurally, and accumulate reusable procedural memory over time.

We therefore study efficient skill use under large-scale skill
libraries. Our goal is to improve task success while reducing token cost,
runtime overhead, and repeated skill organization. To this end, we propose
\model{}, a lightweight structure-aware procedural memory framework. \model{}
maintains two complementary memory layers. Atomic Skill Memory represents
individual executable skills with their semantic, interface, and feedback
signals. Skill-DAG Memory stores reusable skill compositions distilled
from successful executions. Given a new task, \model{} first retrieves relevant
procedural memory and reuses a high-confidence Skill-DAG template when
available. Otherwise, or when the current guidance becomes insufficient,
Skill-GNN constructs a task-conditioned candidate graph. It then performs state-aware skill selection over retrieved atomic skills. After execution,
successful trajectories are compressed into Skill-DAGs and written back to
memory for future reuse.

We evaluate \model{} on SkillsBench and ALFWorld~\cite{li2026skillsbench,shridhar2020alfworld}.
SkillsBench tests large-scale skill retrieval and technical task execution. ALFWorld evaluates long-horizon dependencies, structured skill use, and
cross-task reuse in interactive environments. We compare \model{} with full
skill loading, flat vector retrieval, and graph-based skill selection baselines.

Our contributions are fourfold:
\begin{itemize}
    \item We propose \model{}, a dual-layer procedural memory framework that combines Atomic Skill Memory for fine-grained skill access with Skill-DAG Memory for reusable procedural structures.
    \item We design a Skill-GNN module for state-aware skill selection over task-conditioned candidate graphs.
    \item We introduce trajectory-DAG extraction to distill successful executions into reusable Skill-DAG templates and update procedural memory through execution feedback.
    \item We evaluate \model{} on ALFWorld and SkillsBench, achieving 87.1\% ALFWorld success and 38.38\% SkillsBench reward while reducing token and runtime costs over strong structure-aware baselines.
\end{itemize}

\section{Related Work}

\paragraph{From Tool Calling to Large-Scale Skill Retrieval.}
Early tool-augmented language models focused on tool invocation, argument generation, and reasoning with returned results. Pioneering works like ReAct, Toolformer, and surveys established the paradigm via reasoning-action interleaving, self‑supervised learning, and systematic analysis~\cite{yao2023react,schick2023toolformer,mialon2023augmented}. As external capabilities grow from a few tools to large API or skill libraries, retrieval becomes essential. Gorilla, API‑Bank, ToolBench, ToolLLM, and related methods reduce context overhead and calling errors through documentation retrieval, embedding, hierarchical search, task‑specific construction, or query rewriting~\cite{patil2023gorilla,li2023apibank,xu2023toolbench,qin2024toolllm,hao2023toolkengpt,du2024anytool,yuan2023craft,chen2024reinvoke}. However, most retrieval methods still treat capabilities as independent textual entries. They mainly optimize semantic relevance.
Skill retrieval is harder due to procedural knowledge, execution conditions, input‑output constraints, and task experience. Benchmarks like SkillsBench and SWE‑Skills‑Bench show that excessive or mismatched skills increase context cost and can hurt performance~\cite{li2026skillsbench,han2026sweskillsbench}. Thus, large‑scale skill use demands compact, relevant, and executable selection, not just semantic similarity.

\paragraph{From Structured Skills to Evolving Procedural Memory.}
As skills become reusable assets, recent work organizes them with explicit
structures. SkillNet represents skills through a unified ontology and relation
network. AgentSkillOS manages skills through classification, retrieval,
capability trees, and dynamic skill management~\cite{liang2026skillnet,li2026agentskillos}.
Graph-based methods further model relations among tools or skills. ToolNet
supports graph navigation over large tool collections; Graph of Skills retrieves
dependency-aware skill bundles from a skill dependency graph; and GraSP
constructs DAG-style skill structures with precondition and effect relations for
long-horizon tasks~\cite{liu2024toolnet,liu2026gos,xia2026grasp}. These methods
show that explicit structures help recover dependencies and execution relations. However, many of them use structure mainly for the
current task or as a static organization layer. They give limited attention to how
successful executions should be distilled, updated, and reused as evolving
procedural memory. \model{} instead converts successful trajectories into
reusable Skill-DAG templates and updates their reliability through execution
feedback.

\begin{figure*}[t]
  \centering
  \includegraphics[width=0.9\textwidth]{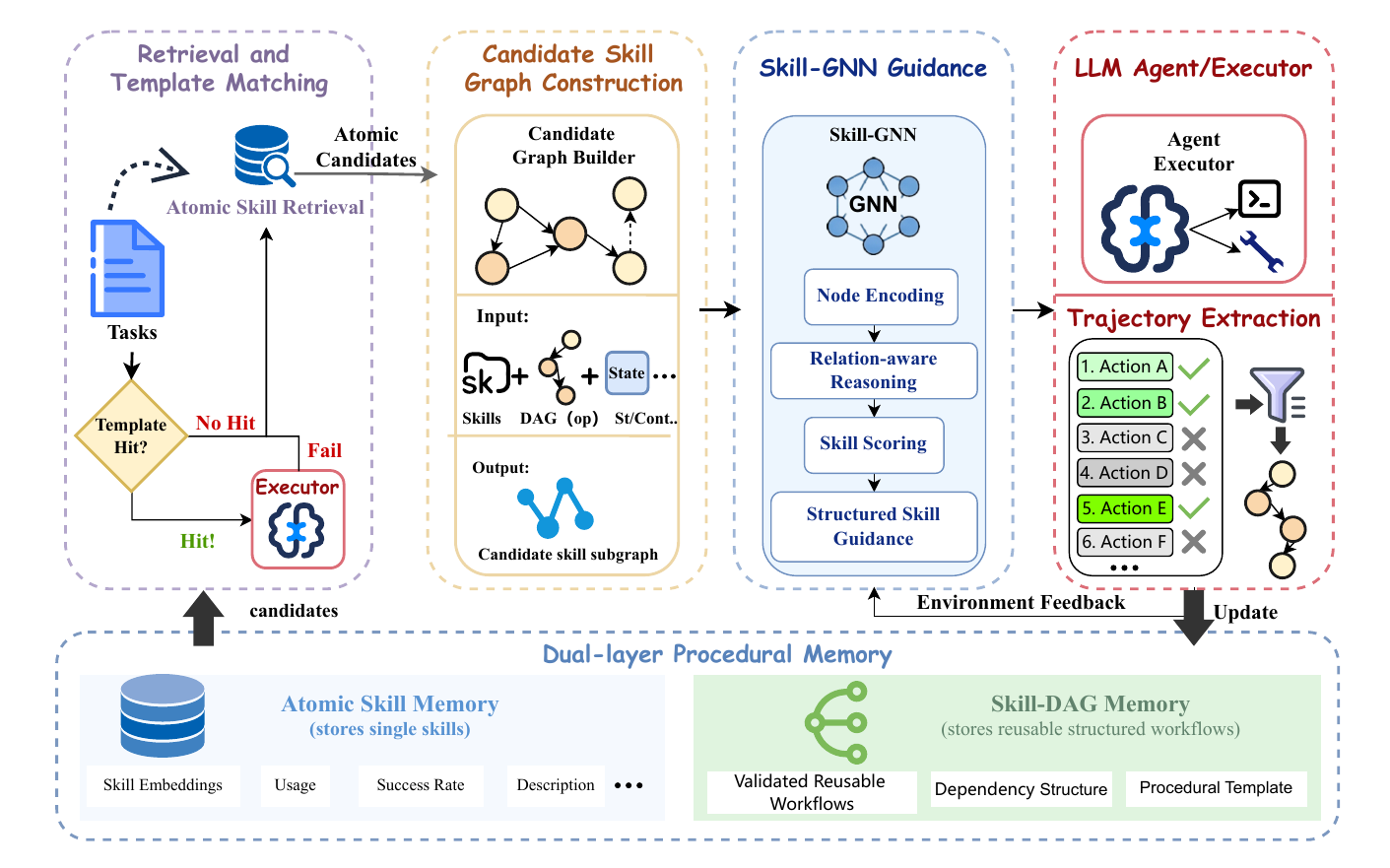} 
  \setlength{\abovecaptionskip}{0pt}
  \caption{Overview of \model{}. The system retrieves atomic skills or matched Skill-DAG templates from dual-layer procedural memory, constructs a task-conditioned candidate skill graph, and uses Skill-GNN to produce compact guidance for the LLM executor. Successful trajectories are distilled back into reusable Skill-DAG memory.}
  \label{framework}
\end{figure*}
  
\paragraph{Lightweight Skill Retrieval and Selection.}
Large-scale skill use imposes constraints on context, latency, and
computation. Recent work explores lightweight skill routing, representation, and
reuse. SkillRouter uses compact models for recall-then-rerank selection over
large skill sets. Formal Skill replaces verbose Markdown descriptions with
executable skill structures. CUA-Skill encapsulates GUI operation knowledge
into reusable skills for dynamic retrieval and parameter
instantiation~\cite{zheng2026skillrouter,zhang2026formalskill,chen2026cuaskill}.
Other methods emphasize reusable experience. ProcMEM converts interaction
experience into procedural memories. SkillDAG maintains typed relations among
skills and evolves across episodes. Compositional Skill Routing retrieves
skills for decomposed subtasks and composes them with a dependency-aware DAG
planner~\cite{mi2026procmem,bai2026skilldag,gao2026compositional}. These studies
show that lightweight skill use is not merely reducing retrieved skills, but
jointly concerns compact representation, efficient selection, execution
constraints, and experience reuse. \model{} addresses this gap through a unified
procedural memory framework that combines bounded atomic retrieval, reusable
Skill-DAG Memory, and Skill-GNN organization.

  \begin{figure*}[t]
      \centering
      \includegraphics[width=\textwidth]{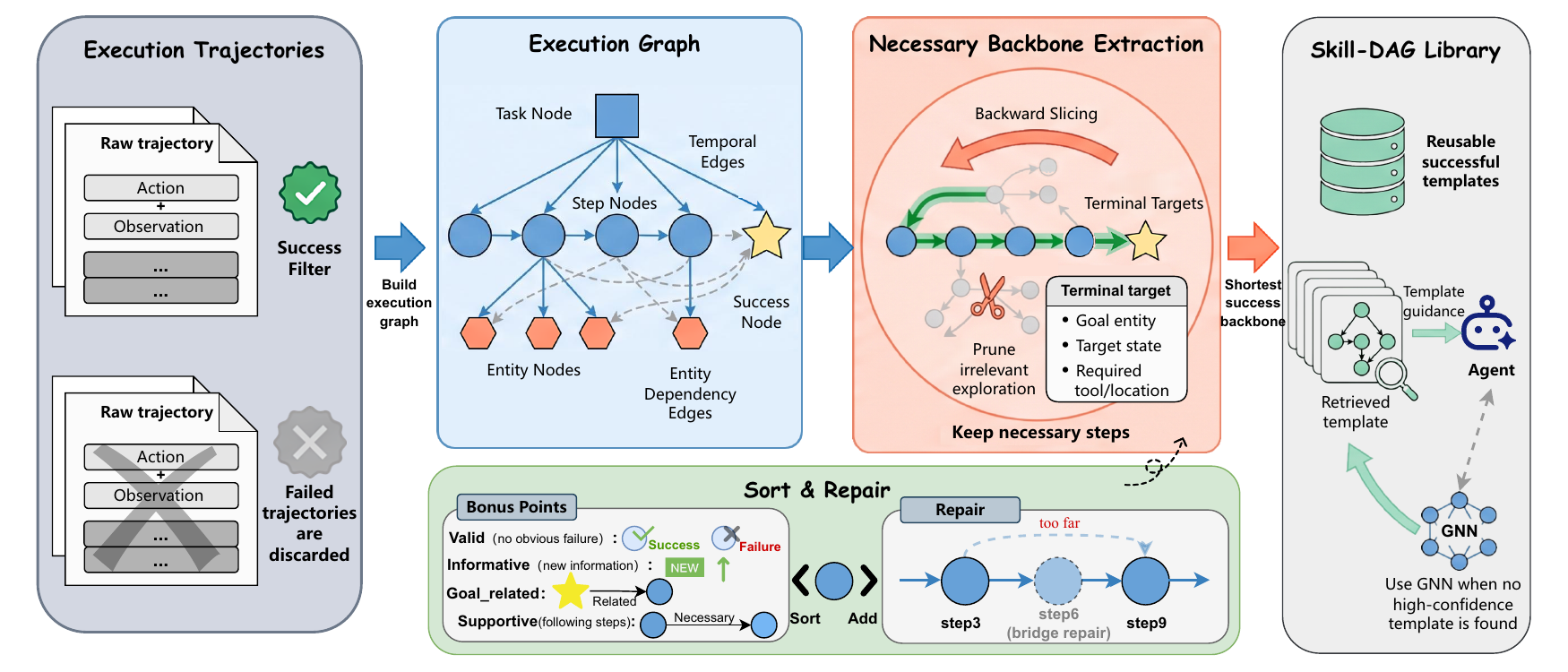}
      \setlength{\abovecaptionskip}{0pt}
      \caption{Overview of trajectory-DAG extraction. Successful trajectories are converted into execution graphs, from which necessary steps are identified through backward slicing, pruning, and dependency repair. The resulting backbones are stored as reusable Skill-DAG templates for future tasks.}
      \label{fig:trajectory_dag}
  \end{figure*}

\section{Method}

Given a task request and a large-scale skill library, \model{} provides compact,
executable skill guidance through two memory layers: Atomic Skill Memory for
individual skills and Skill-DAG Memory for reusable compositions distilled from
successful executions. As shown in Figure~\ref{framework}, \model{} reuses a
matched Skill-DAG template when available. Otherwise it retrieves atomic skills and invokes
Skill-GNN for state-aware reranking when needed. Successful executions are
written back as new Skill-DAGs. This enables procedural memory to accumulate over
time. The full protocol is provided in Section~A.1 of the supplementary material.

\subsection{Dual-Layer Procedural Memory}

\model{} separates procedural knowledge into two complementary memory layers. Atomic Skill Memory supports fine-grained access to individual skills, while Skill-DAG Memory supports direct reuse of successful skill compositions. The supplementary material provides the full schema and update rules in Section~A.2.

\paragraph{Atomic Skill Memory.}
Atomic Skill Memory represents each skill as a reusable executable unit rather
than an isolated text snippet. We denote it as
\begin{equation}
\mathcal{M}_{a}=\{s_i\}_{i=1}^{N},
\end{equation}
where each skill is represented as
\begin{equation}
s_i=(d_i,e_i,\sigma_i).
\end{equation}
Here, $d_i$ denotes skill metadata, $e_i$ denotes the semantic embedding.
$\sigma_i$ denotes lightweight execution feedback, such as usage frequency,
success rate, and execution cost. This allows retrieval to consider both semantic relevance and historical execution evidence without loading the full
skill library into the LLM context.

\paragraph{Skill-DAG Memory.}
Skill-DAG Memory stores reusable procedural templates extracted from successful
executions. We denote it as
\begin{equation}
\mathcal{M}_{d}=\{G_j\}_{j=1}^{M},
\end{equation}
where each Skill-DAG is represented as
\begin{equation}
G_j=(V_j,E_j,e_j,m_j).
\end{equation}
Here, $V_j$ contains procedural nodes, $E_j$ contains directed dependency edges,
$e_j$ is the task embedding associated with the template, and $m_j$ stores
compact execution metadata. A Skill-DAG captures a reusable execution structure. It allows the agent to reuse previously successful procedures instead of reconstructing them from scratch.

\paragraph{Procedural Memory Retrieval.}
Given a task request $q$, \model{} first searches Skill-DAG Memory for reusable templates:
\begin{equation}
\mathcal{C}_{d}(q)=
\{G_j \in \mathcal{M}_{d}\mid \operatorname{sim}(e_q,e_j)\geq \tau\},
\end{equation}
where $e_q$ is the task embedding and $\tau$ is the template matching threshold. If $\mathcal{C}_{d}(q)$ is non-empty, the highest-scoring Skill-DAG is injected as structured procedural guidance. If no template satisfies the threshold, \model{} retrieves a bounded set of atomic candidates:
\begin{equation}
\mathcal{C}_{a}(q)=
\operatorname{TopK}_{s_i\in\mathcal{M}_{a}}
\operatorname{sim}(e_q,e_i).
\end{equation}
The retrieved template or atomic candidates are then passed to the subsequent Skill-GNN orchestration stage when structure-aware organization is needed. This retrieval strategy avoids full-library loading while preserving two complementary modes of reuse: direct reuse of successful Skill-DAG templates and flexible organization of atomic skills for novel task states.

\subsection{Structure-Aware Skill Selection with Skill-GNN}

When reusable templates are unavailable or insufficient for the current execution context, \model{} uses Skill-GNN to perform structure-aware ranking and selection over retrieved atomic candidates. The goal is to identify semantically relevant skills. It also aims to select skills that are compatible with the task state and dependency relations among candidates.

\paragraph{Candidate Skill Graph Construction.}
Given a task request $q$, retrieved atomic candidates $\mathcal{C}_{a}(q)$, and
the current execution context, \model{} constructs a task-conditioned candidate
graph
\begin{equation}
\mathcal{G}_{q}=(\mathcal{V}_{q},\mathcal{E}_{q},\mathcal{R}),
\end{equation}
where $\mathcal{V}_{q}$ contains task, context, and candidate skill nodes.
$\mathcal{E}_{q}$ contains typed edges, and $\mathcal{R}$ denotes relation
types. Skill-node features combine semantic embeddings with lightweight
execution feedback. Task and context nodes encode the request, current
state, and execution history. Edges connect task--skill, context--skill, and
skill--skill pairs, capturing relevance, state compatibility, semantic
relatedness, and available dependency cues. Additional construction details are
provided in Section~A.5 of the
supplementary material.

\paragraph{Skill-GNN Scoring.}
Given $\mathcal{G}_{q}$, Skill-GNN first performs node encoding and relation-aware reasoning through message passing, as summarized in Figure~\ref{framework}. Let $h_i^{(0)}$ be the initial representation of node $i$. At layer $\ell$, the node representation is updated by aggregating messages from relation-specific neighbors:
\begin{equation}
h_i^{(\ell+1)} =
\phi_{\ell}\left(
h_i^{(\ell)},
\sum_{r \in \mathcal{R}}
\sum_{j \in \mathcal{N}_{r}(i)}
\alpha_{ij}^{r} W_{r}^{(\ell)} h_j^{(\ell)}
\right),
\end{equation}
where $\mathcal{N}_{r}(i)$ denotes neighbors of node $i$ under relation type $r$, $W_{r}^{(\ell)}$ is a relation-specific transformation. $\alpha_{ij}^{r}$ is the message weight, and $\phi_{\ell}$ is the layer update function.

After $L$ message-passing layers, Skill-GNN performs skill scoring. It assigns each candidate skill $s_i \in \mathcal{C}_{a}(q)$ a context-dependent utility score:
\begin{equation}
p_i=f_{\theta}\left(h_i^{(L)}\right).
\end{equation}
The top-ranked skills, together with compact dependency cues when available, are injected as structured guidance for the LLM agent. This graph-based scoring incorporates task relevance, runtime state, and inter-skill dependencies into the skill selection process.

\paragraph{Cold-Start Deployment.}
\model{} supports both trajectory-rich and trajectory-scarce deployment settings.
When target-domain trajectories are available, Skill-GNN is trained with
trajectory-prefix supervision, where each prefix defines the current procedural
state and the next executed skill is used as the target. When trajectories are
scarce or unavailable, \model{} uses a library-level warm start constructed only
from the skill library:
\begin{align}
\widetilde{\mathcal{D}}_{0}
&=
\{(\tilde{q}_{i}, s_i)\}_{s_i\in\mathcal{M}_{a}},\\
\widetilde{\mathcal{E}}_{0}
&=
\{(s_i,r,s_j)\mid \kappa_r(s_i,s_j)\ge \eta_r\},
\end{align}
where $\tilde{q}_{i}$ is a pseudo query generated from the metadata of skill
$s_i$, and $\widetilde{\mathcal{E}}_{0}$ contains weak skill--skill relations
induced by semantic similarity, shared tools, tags, or schema compatibility.
This initialization uses no task outcomes, verifier rewards, or evaluation
trajectories. During early deployment, Atomic Skill Memory retrieval and
available Skill-DAG templates provide the primary guidance, while the
warm-started Skill-GNN is used only for bounded reranking. As successful
executions accumulate, trajectory-DAG extraction converts them into new
Skill-DAG templates and trajectory-prefix training samples, gradually adapting
the system from retrieval-only guidance to state-aware graph selection. Since
Skill-GNN is inductive over dynamically constructed candidate graphs, it can
generalize to unseen tasks, unseen skill combinations, and newly extracted
Skill-DAG variants. Additional protocol and leakage-prevention details are
provided in Section~A.4 of the supplementary material.

\subsection{Trajectory-DAG Extraction}
As shown in Figure~\ref{fig:trajectory_dag}, \model{} converts successful interaction trajectories into reusable Skill-DAG templates after execution. Instead of storing complete trajectories, which may contain failed attempts, redundant exploration, or environment-specific details, the system extracts a compact execution backbone.  This backbone preserves the necessary steps for reaching the success condition. Given an execution trajectory $\tau$, the system first constructs an execution graph
\begin{equation}
G_{\tau}=(V_{\tau},E_{\tau}),
\end{equation}
where nodes represent executed steps and task-relevant states. Edges encode temporal order, entity dependencies, state transitions, and terminal reachability.

Starting from the successful terminal node $v_{\mathrm{succ}}$, the system performs backward slicing to identify success-relevant nodes:
\begin{equation}
V^{*}_{\tau}=\{v \in V_{\tau} \mid v \to^{*} v_{\mathrm{succ}}\},
\end{equation}
where $v \to^{*} v_{\mathrm{succ}}$ indicates that $v$ reaches the successful terminal state through temporal or dependency paths. Failed branches, redundant exploration, and operations that do not contribute to the final outcome are removed. The remaining nodes and their induced dependency edges are ordered and repaired according to temporal and dependency constraints, forming a compact Skill-DAG:
\begin{equation}
G^{*}_{\tau}=(V^{*}_{\tau},E^{*}_{\tau}).
\end{equation}
Finally, $G^{*}_{\tau}$ is written back to Skill-DAG Memory for future reuse. Successful non-template executions introduce new templates; successful template-guided executions reinforce or refine existing templates. Failed template executions only update reliability statistics. Detailed extraction rules, scoring signals, and update policies are provided in Section~A.3 of the supplementary material.

\section{Experiments}
\begin{table*}[t]
\centering
\small
\renewcommand{\arraystretch}{1.12}
\setlength{\tabcolsep}{5pt}
\begin{tabular}{llccc|cccc}
\toprule
\multirow{2}{*}{\textbf{Backbone}} &
\multirow{2}{*}{\textbf{Method}}
& \multicolumn{3}{c|}{\textbf{SkillsBench}}
& \multicolumn{4}{c}{\textbf{ALFWorld}}\\
\cmidrule(lr){3-5} \cmidrule(lr){6-9}
& & R $\uparrow$ & T $\downarrow$ & S $\downarrow$
& Success $\uparrow$ & Steps $\downarrow$ & Tokens $\downarrow$ & Runtime $\downarrow$ \\
\midrule

\multirow{5}{*}{\textbf{MiniMax-M2.7}}
& Vanilla Skills
& 21.61 & 798,323 & 469.3
& 59.3 & 21.24 & 152,242.1 & 55.95s \\
& Vector Skills
& 13.07 & \underline{722,765} & 439.5
& 63.9 & 18.36 & \underline{55,816.6} & 55.71s \\
& GoS
& 23.50 & 735,113 & \underline{399.4}
& \underline{68.6} & 20.26 & 62,033.9 & \underline{54.22s} \\
& SkillDAG
& \textbf{27.09} & 979,781 & 476.1
& 63.3 & \textbf{16.37} & 296,512.4 & 137.38s \\
& \textbf{Ours}
& \underline{25.14} & \textbf{721,926} & \textbf{360.5}
& \textbf{70.7} & \underline{18.29} & \textbf{45,785.7} & \textbf{51.56s} \\

\midrule

\multirow{5}{*}{\textbf{GPT-5.2-Codex}}
& Vanilla Skills
& 28.87 & 2,785,485 & \underline{686.8}
& 84.9 & \underline{14.65} & 99,023.9 & 84.57s \\
& Vector Skills
& 23.44 & \underline{1,047,698} & 763.4
& 83.7 & 16.18 & \textbf{47,357.3} & 80.43s \\
& GoS
& \underline{37.50} & 1,162,375 & 705.2
& 85.3 & 16.46 & 48,805.6 & \underline{71.23s} \\
& SkillDAG
& 34.38 & 1,370,270 & 861.1
& \underline{85.7} &\textbf{ 11.73} & 105,449.9 & 169.10s \\
& \textbf{Ours}
& \textbf{38.38} & \textbf{980,029} & \textbf{618.7}
& \textbf{87.1} & 15.07 & \underline{47,409.1} & \textbf{66.01s} \\

\bottomrule
\end{tabular}
\caption{Main results on SkillsBench and ALFWorld under different backbone models.
For SkillsBench, R denotes the verifier reward, T denotes average input tokens,
and S denotes agent-only runtime in seconds. For ALFWorld, Success denotes task
success rate, while Steps, Tokens, and Runtime denote average environment steps,
average token consumption, and agent-only runtime, respectively.
$\uparrow$ indicates higher is better, and $\downarrow$ indicates lower is better.
Bold numbers indicate the best result within each backbone group and benchmark,
and underlined numbers indicate the second-best result.}
\label{tab:main_results}
\end{table*}
\subsection{Experimental Setup}

\paragraph{Benchmarks.}
We evaluate \model{} on two complementary skill-use benchmarks.
\textbf{ALFWorld} \cite{shridhar2020alfworld} is an interactive text-based embodied environment. It requires multi-step household task completion, testing long-horizon dependencies, state transitions, and procedural reuse.
\textbf{SkillsBench} \cite{li2026skillsbench} evaluates technical agent tasks with large skill libraries. It tests whether agents can retrieve and apply task-relevant skills under limited context budgets.

\paragraph{Metrics.}
For ALFWorld, we report Success, Steps, Tokens, and Runtime.
Success ($\uparrow$) is the task success rate within the maximum step budget. Steps ($\downarrow$) is the average number of environment interactions. Tokens ($\downarrow$) is the average token consumption during agent execution, and Runtime ($\downarrow$) is the average agent-only execution time.

For SkillsBench, we report verifier reward $R$ ($\uparrow$), average input tokens $T$ ($\downarrow$), and agent-only runtime $S$ ($\downarrow$).
Following the benchmark protocol, $R$ is determined by the official verifier.
For fair accounting, $S$ measures the agent execution phase and excludes environment setup, dependency installation, Docker build time, and verifier runtime. For \model{}, online skill retrieval or reranking invoked during execution is included in the measured agent phase.

\paragraph{Baselines.}
We compare \model{} with four baselines that form a controlled progression in
skill access and structure awareness. 
\textbf{Vanilla Skills} loads the full
skill library, testing whether exhaustive context alone is sufficient.
\textbf{Vector Skills} retrieves top-$k$ skills by embedding similarity,
isolating the effect of flat semantic retrieval.
\textbf{GoS} serves as a
representative static graph-retrieval method, using a prebuilt skill graph to
retrieve structurally related skills. 
\textbf{SkillDAG} serves as a recent
sota baseline for dependency-aware skill selection with
a self-evolving graph structure.
\textbf{\model{} (Ours)} further introduces
dual-layer procedural memory, trajectory-DAG extraction, and on-demand
Skill-GNN reranking. These baselines allow us to distinguish gains from more
context, semantic retrieval, static graph structure, typed graph selection, and
reusable procedural memory.

\paragraph{Implementation Details.}
For fair comparison, all methods use the same benchmark splits, skill libraries, agent interface, execution environment, and verification protocol.
We evaluate with two backbone LLMs, MiniMax-M2.7 \cite{minimax2026m27} and GPT-5.2-Codex \cite{openai2025gpt52codex}, representing weaker and stronger model settings.

On \textbf{ALFWorld}, we use the dev split with 140 episodes and set the maximum number of environment steps to 30.
All skill-based methods use the same library of 37 \texttt{alfworld-*} skills.
\model{} builds Atomic Skill Memory from this library, constructs Skill-DAG Memory only from successful non-evaluation trajectories, and trains Skill-GNN with trajectory-prefix supervision under disjoint training and evaluation instances.

On \textbf{SkillsBench}, we follow the GoS execution setting with Docker-based environments, Harbor agent execution, verifier-based evaluation, at most two attempts per task, and two concurrent trials.
Following the benchmark convention, tasks that remain unresolved due to
infrastructure failures after the allowed retries are excluded rather than
counted as model failures, leaving the same 67 valid model tasks for all methods.
All methods use the same \texttt{skills\_200} skill library.
\model{} builds Atomic Skill Memory from the same library and applies library-level warm start plus Skill-GNN reranking.
Detailed hyperparameters, leakage-prevention rules, retry handling, and additional inference details are provided in Section~B of the supplementary material.

\begin{figure*}[t]
  \centering
  \includegraphics[width=0.75\textwidth]{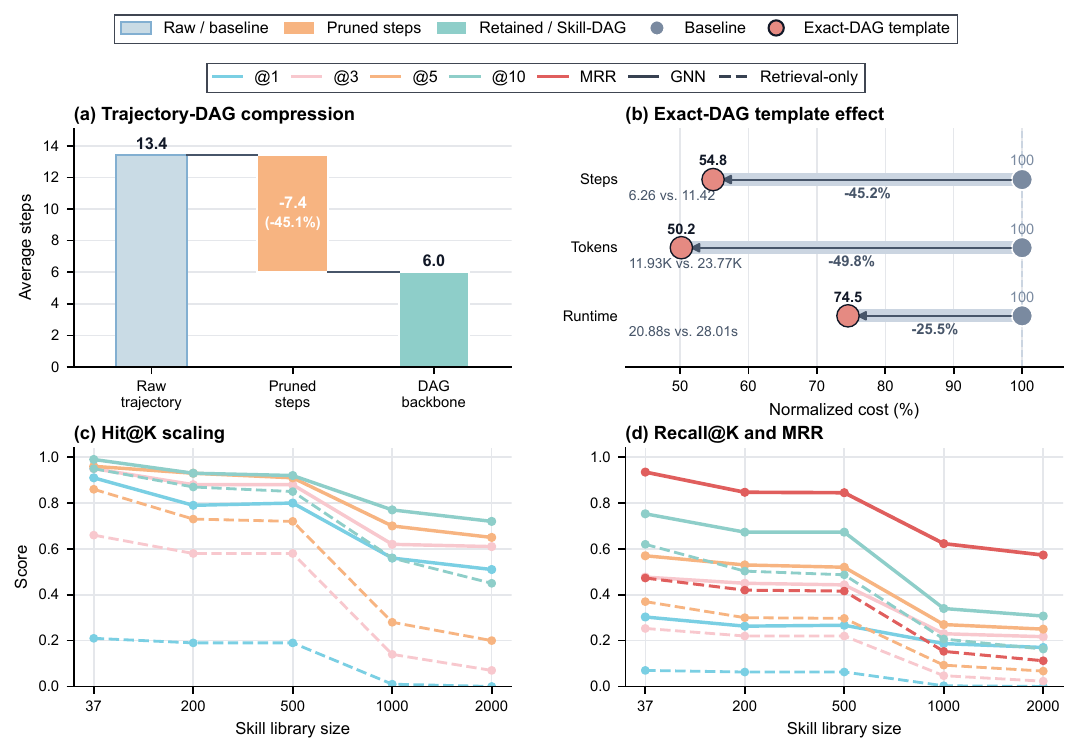}
  \setlength{\abovecaptionskip}{0pt}
  \caption{Effectiveness and scalability analysis.
(a) Trajectory-DAG extraction compresses successful trajectories into compact DAG backbones.
(b) Exact-DAG template reuse reduces normalized execution cost.
(c) Hit@K under different skill library sizes.
(d) Recall@K and MRR under different skill library sizes. Solid lines denote GNN-based reranking, while dashed lines denote retrieval-only selection.}
\label{fig:dag_scaling_analysis}
\end{figure*}
\subsection{Main Results}

\paragraph{Overall effectiveness--efficiency trade-off.}
Table~\ref{tab:main_results} shows that \model{} achieves the best overall
trade-off across both benchmarks and backbone models. On ALFWorld, \model{}
obtains the highest success rate under both MiniMax-M2.7 and GPT-5.2-Codex,
improving over GoS by 2.1 and 1.8 percentage points, respectively, while also
achieving the lowest runtime. On SkillsBench, \model{} achieves the best reward
with GPT-5.2-Codex and the second-best reward with MiniMax-M2.7, where SkillDAG
obtains a higher reward but requires substantially more tokens and runtime. In
particular, compared with SkillDAG under MiniMax-M2.7, \model{} reduces input
tokens by 26.3\% and runtime by 24.3\%, while remaining competitive in reward.

\paragraph{Efficiency becomes more important with stronger models.}
As the backbone model becomes stronger, structure-aware baselines such as GoS
and SkillDAG also obtain competitive rewards, especially on SkillsBench. This
makes efficiency a critical differentiator rather than a secondary metric.
Under GPT-5.2-Codex, \model{} improves SkillsBench reward over GoS from
37.50\% to 38.38\%, while reducing input tokens by 15.7\% and runtime by
12.3\%. On ALFWorld, SkillDAG achieves fewer steps, but it incurs much higher
token and runtime costs. In contrast, \model{} maintains the highest success
rate with much lower execution overhead, suggesting that reusable procedural
memory is most useful when it provides compact guidance rather than longer
plans or larger context.

\paragraph{Why structure matters.}
The comparison among Vector Skills, GoS, SkillDAG, and \model{} clarifies the
role of structure. Vector Skills often reduce context size but can lose
important procedural dependencies, leading to weaker reward or success. GoS and
SkillDAG show that explicit skill structures improve task performance, but they
mainly rely on static or pre-constructed structures and may introduce extra
context or execution overhead. \model{} further combines Atomic Skill Memory,
reusable Skill-DAG Memory, and state-aware Skill-GNN reranking, allowing the
agent to reuse successful procedural structures while keeping injected guidance
compact. These results indicate that the main advantage of \model{} is not
simply retrieving more skills, but selecting and reusing structured procedural
guidance under limited token and runtime budgets.

Detailed case analyzes of \model{} are provided in Section~D
of the supplementary material.

\begin{table}[t]
\centering
\small
\setlength{\tabcolsep}{4pt}
\renewcommand{\arraystretch}{1.08}
\begin{tabular}{lcccc}
\toprule
\textbf{Backbone} & \textbf{Hit} & \textbf{Fallback} & \textbf{Ineffective} & \textbf{Hit Success} \\
\midrule
\textbf{MiniMax-M2.7} & 56.43 & 43.57 & 25.32 & 74.68 \\
\textbf{GPT-5.2-Codex} & 60.71 & 39.29 & 14.12 & 85.88 \\
\bottomrule
\end{tabular}
\caption{Template-availability analysis on ALFWorld. All values are percentages.}
\label{tab:template_availability}
\end{table}
\subsection{Analysis of DAG-Template Effectiveness}
We analyze whether the extracted Skill-DAGs are both compact and useful.
As shown in Figure~\ref{fig:dag_scaling_analysis}(a), trajectory-DAG extraction
compresses successful trajectories from 13.4 average steps to 6.0 backbone
steps, pruning 54.9\% of operations while preserving the executable core.
This preliminarily demonstrates the compression effect of DAG templates.
We then evaluate the availability and effectiveness of DAG templates.
As shown in Table~\ref{tab:template_availability}, on ALFWorld, \model{}
achieves template hit rates of 56.43\% and 60.71\% under MiniMax-M2.7 and
GPT-5.2-Codex, respectively. Among template-hit tasks, the success rates are
74.68\% and 85.88\%, respectively. These results indicate that DAG templates can
accumulate reusable procedural structures for later tasks. Unlike ALFWorld,
SkillsBench contains highly heterogeneous technical tasks, where exact
procedural-template reuse is naturally limited. Therefore, we mainly use
SkillsBench to evaluate large-scale atomic-skill retrieval and structure-aware
reranking, and use ALFWorld to analyze Skill-DAG template reuse.
Finally, we conduct an exact-template analysis on ALFWorld. This isolates the effect of a correct DAG skeleton itself. For each successfully completed task, we
extract its trajectory and directly provide the DAG when re-executing
that task. This setting tests whether DAG templates can improve execution
efficiency. As shown in Figure~\ref{fig:dag_scaling_analysis}(b), exact-DAG
guidance preserves success on all evaluated tasks while reducing average steps
from 11.42 to 6.26, token consumption from 23.77K to 11.93K, and runtime from
28.01s to 20.88s, corresponding to reductions of 45.2\%, 49.8\%, and 25.5\%,
respectively. This suggests that templates provide effective procedural guidance
for improving task efficiency.


\begin{table}[htbp]
\centering
\small
\renewcommand{\arraystretch}{1.12}
\setlength{\tabcolsep}{3.8pt}
\begin{tabular}{lcccc}
\toprule
Method Variant & Succ. $\uparrow$ & Steps $\downarrow$ & Time $\downarrow$ & Tokens $\downarrow$ \\
\midrule
w/o Skill-DAG  & 56.8 & 23.0 & 55.4 & 67.5K \\
w/o Skill-GNN$^\dagger$ & 58.3 & 20.3 & 54.7 & 53.6K \\
w/o DAG Extraction & 63.0 & 18.9 & 57.8 & 56.2K \\
Full \model{} & \textbf{70.7} & \textbf{18.3} & \textbf{51.6} & \textbf{45.8K} \\
\bottomrule
\end{tabular}
\caption{Component ablation of \model{} on ALFWorld. Succ., Steps, Time, and Tokens denote task success rate, average completion steps, average runtime, and average token consumption. $^\dagger$ removes graph-based Skill-GNN and uses retrieved procedural memory.}
\label{tab:ablation}
\end{table}
\subsection{Ablation Study.}
\paragraph{\model{} Component Analysis}
Table~\ref{tab:ablation} evaluates the contribution of each component of \model{} on ALFWorld. 
\begin{itemize}
  \item \textbf{Skill-DAG templates} provide reusable procedural guidance. Removing them reduces success to 56.8\% and increases average steps to 23.0, indicating more redundant exploration. 
  \item \textbf{Skill-GNN} supports state-aware skill selection: without graph-based scoring, success drops to 58.3\%, showing that retrieved memory alone is not sufficient for stable execution under changing task states. 
  \item \textbf{DAG extraction} makes procedural memory compact. Replacing extracted Skill-DAG backbones with complete execution traces lowers success to 63.0\%. It also increases token and runtime cost, because full traces contain redundant exploration and environment-specific details. 
  \item \textbf{The full \model{}} achieves the best results across all metrics. This demonstrates that reusable Skill-DAG templates, structure-aware skill selection, and compact trajectory extraction are complementary.
\end{itemize}

\paragraph{Scalability to Larger Skill Libraries.}
Figures~\ref{fig:dag_scaling_analysis}(c) and (d) further examine whether structure-aware skill selection remains effective as the skill library grows. Retrieval-only selection degrades rapidly when the library size increases, especially at 1000 and 2000 skills, where semantic similarity alone struggles to rank the required skills near the top. In contrast, Skill-GNN consistently achieves higher Hit@K, Recall@K, and MRR across all library sizes. For example, at 2000 skills, Skill-GNN improves Hit@1 from 0.0 to 0.51 and MRR from 0.112 to 0.573 over retrieval-only selection. Although both methods become harder to rank as the candidate pool expands, the gap remains substantial, showing that graph-based contextualization helps preserve reliable skill selection under large-scale skill libraries.

Additional backbone comparisons and hyperparameter sensitivity analyses can be found in Section~C of the supplementary material.

\section{Conclusion}

We presented \model{}, a reusable and structure-aware skill memory framework for
LLM-based agents. Instead of treating skills as a flat prompt collection, \model{}
organizes them into Atomic Skill Memory for individual capabilities and
Skill-DAG Memory for reusable procedural structures. Skill-GNN further enables
state-conditioned skill selection over dynamically constructed candidate graphs, allowing the agent to adapt its guidance during execution while remaining inductive to new tasks and skill combinations.
Experiments on ALFWorld and SkillsBench show that \model{} improves both task effectiveness and execution efficiency across weaker and stronger backbone models. The results suggest that effective skill use requires more than semantic retrieval. Agents benefit from compact procedural guidance that captures both which skills are relevant and how they should be composed. Although current performance still depends on retrieval quality and the agent's ability to execute retrieved guidance, \model{} provides a practical step toward scalable, adaptive, and reusable skill memory for LLM-based agents.

\bibliography{aaai2027}


\end{document}


\maketitle
\appendix
\setcounter{section}{0}
\setcounter{subsection}{0}
\setcounter{subsubsection}{0}

\renewcommand{\thesection}{\Alph{section}}
\renewcommand{\thesubsection}{\Alph{section}.\arabic{subsection}}
\renewcommand{\thesubsubsection}{\Alph{section}.\arabic{subsection}.\arabic{subsubsection}}

\begin{abstract}
This supplementary material provides additional implementation details,
experimental results, sensitivity analyses, and case studies for the main paper.
\end{abstract}

\section{Method Details}
\subsection{Overall Inference Protocol}
\label{app:overall_inference}

Algorithm~\ref{alg:overall_inference} summarizes the inference protocol of
\model{}. Given a task request and the current execution state, the system first
retrieves relevant atomic skills and previously extracted Skill-DAG templates.
If a confident template is matched, \model{} injects it as structured procedural
guidance. Otherwise, it falls back to a bounded set of retrieved atomic skills.
During execution, Skill-GNN can be invoked on demand to rerank candidate skills
under the current task, state, and execution history. After the episode
terminates, successful trajectories are converted into Skill-DAG templates and
written back to memory, while failed executions only update reliability
statistics for the used skills or templates.

\begin{algorithm}[h]
\caption{Overall Inference of \model{}}
\label{alg:overall_inference}
\begin{algorithmic}[1]
\REQUIRE task request $q$, initial state $s_0$, Atomic Skill Memory $\mathcal{M}_a$, Skill-DAG Memory $\mathcal{M}_d$
\ENSURE execution result and updated memory

\STATE Retrieve atomic candidates $\mathcal{C}_a$ from $\mathcal{M}_a$
\STATE Retrieve Skill-DAG templates $\mathcal{C}_d$ from $\mathcal{M}_d$

\IF{a confident template exists in $\mathcal{C}_d$}
    \STATE Inject the best Skill-DAG template as procedural guidance
\ELSE
    \STATE Inject top atomic candidates from $\mathcal{C}_a$
\ENDIF

\STATE Initialize trajectory $\tau \leftarrow \emptyset$ and state $s \leftarrow s_0$

\WHILE{the task is not terminated}
    \IF{replanning is triggered}
        \STATE Construct a task-conditioned candidate graph $G(q,s,\tau)$
        \STATE Rerank candidate skills with Skill-GNN
        \STATE Inject top-ranked skills as state-aware guidance
    \ENDIF
    \STATE Agent executes the next action under current guidance
    \STATE Observe new state $s'$ and append the transition to $\tau$
    \STATE $s \leftarrow s'$
\ENDWHILE

\IF{the task succeeds}
    \STATE Extract a Skill-DAG $G_\tau$ from $\tau$
    \STATE Insert or reinforce $G_\tau$ in $\mathcal{M}_d$
\ELSE
    \STATE Update reliability statistics of used skills/templates
\ENDIF
\STATE \RETURN execution result
\end{algorithmic}
\end{algorithm}

\subsection{Memory Schemas}
\label{app:memory_schemas}

\model{} maintains two complementary memory stores: Atomic Skill Memory and
Skill-DAG Memory. Table~\ref{tab:memory-schema} summarizes the main fields and
their roles. Atomic Skill Memory stores individual reusable capabilities, while
Skill-DAG Memory stores procedural templates extracted from successful
executions.

For Atomic Skill Memory, each atomic skill contains three categories of
information. The first category consists of static identity and semantic fields,
which describe what the skill is designed to do. These fields include a stable
identifier, skill name, source path, version, natural-language description, and
a short task-facing capability summary. In our implementation, the retrieval
embedding is computed from the skill name and the one-line capability summary,
so first-stage retrieval focuses on matching the current task to the skill's
task-facing functionality; the full skill document is preserved for prompt
construction after retrieval. The second category consists of interface fields,
which describe how the skill can be invoked, including input and output schemas,
software or tool dependencies, and device or environment requirements. These
fields help the agent use the skill correctly and provide structured signals
for candidate graph construction and compatibility reasoning between neighboring
skills. The third category consists of lightweight dynamic statistics collected
during online use, including usage count, empirical success rate, execution
cost, last-used timestamp, and user preference score. These statistics are
updated after execution and can be used for reranking or personalization. They
are conservatively initialized and updated with an exponential moving average,
allowing the memory to adapt to deployment feedback without changing the
underlying skill definition.

A Skill-DAG is a directed acyclic graph whose nodes correspond to mapped atomic
skills or abstract procedural phases, and whose edges encode typed dependencies
such as temporal, data-flow, control-flow, and argument dependencies. Each
Skill-DAG also stores the source task description, a task embedding for template
retrieval, provenance metadata, dynamic reliability statistics, and a structural
signature for deduplication and merging. During inference, atomic skills and
Skill-DAG templates are retrieved by cosine similarity in their respective
embedding spaces. After execution, only successful trajectories can insert new
Skill-DAG templates; failed executions update reliability statistics but do not
create positive procedural memory.

\begin{table*}[t]
\centering
\small
\begin{tabularx}{\textwidth}{>{\centering\arraybackslash}p{0.10\textwidth} p{0.12\textwidth} p{0.34\textwidth} X}
\toprule
Memory & Sub-type & Key Fields & Purpose \\
\midrule
\multirow{Atomic Skill} & Core & \texttt{skill\_id}\newline\texttt{name}\newline\texttt{one\_line\_capability}\newline\texttt{description}\newline\texttt{semantic\_embedding} & Represent individual reusable capabilities and support task-skill retrieval. \\
\cmidrule{2-4}
 & Interface & \texttt{input\_schema}\newline\texttt{output\_schema}\newline\texttt{tooling}\newline\texttt{device\_requirements}\newline\texttt{raw\_content} & Describe how a skill can be invoked and provide full skill content for prompt injection. \\
\cmidrule{2-4}
 & Statistics & \texttt{success\_rate}\newline\texttt{exec\_cost}\newline\texttt{usage\_count}\newline\texttt{last\_used\_at} & Track online reliability and cost for reranking or adaptation. \\
\midrule
\multirow{DAG} & Template & \texttt{dag\_id}\newline\texttt{task\_description}\newline\texttt{task\_embedding}\newline\texttt{nodes}\newline\texttt{edges} & Store reusable multi-step procedural structures extracted from successful executions. \\
\cmidrule{2-4}
 & Node/Edge & \texttt{skill\_id}\newline\texttt{action\_type}\newline\texttt{arg\_tokens}\newline\texttt{edge\_type}\newline\texttt{dep\_source}\newline\texttt{weight} & Represent skill bindings, procedural phases, and typed dependencies between steps. \\
\cmidrule{2-4}
 & Metadata & \texttt{meta}\newline\texttt{structural\_signature}\newline\texttt{success\_rate}\newline\texttt{usage\_count}\newline\texttt{parent\_dag\_ids} & Support provenance tracking, deduplication, merging, and reliability-aware reuse. \\
\bottomrule
\end{tabularx}
\caption{Compact schema of the two memory stores used by \model{}.}
\label{tab:memory-schema}
\end{table*}

\subsection{Trajectory-DAG Extraction}
\label{app:trajectory_dag_extraction}

Trajectory-DAG extraction converts successful executions into reusable
procedural templates. It complements the memory update step in
Algorithm~\ref{alg:overall_inference} by specifying how an interaction trace is
compressed into a Skill-DAG. We first normalize environment-specific logs into
an ordered trajectory
\[
\tau=\{(a_t,o_t,r_t,y_t,b_t)\}_{t=1}^{T},
\]
where $a_t$ and $o_t$ denote the action and observation, $r_t$ and $y_t$ are
optional reward and completion signals, and $b_t$ is an optional runtime skill
binding.

For each normalized step, the extractor parses the action into a lightweight
schema containing the operation verb, argument spans, and mentioned entities.
It then builds an intermediate execution graph with step nodes and typed
dependencies: temporal edges connect adjacent useful steps, entity-dependency
edges connect steps sharing task-relevant entities or arguments, and terminal
edges connect completion-related steps to the success node.

To avoid storing noisy exploration, the extractor keeps only a compact execution
backbone. It seeds the retained set with terminal or positive-reward steps and
scans the trajectory backward to preserve prerequisite steps that introduce
required entities, satisfy argument dependencies, or directly support retained
steps. Low-information no-op steps, explicit failures, and redundant exploration
are removed. In implementation, each step receives an interpretable usefulness
score:
\[
u_t=\mathrm{clip}_{[0,1]}(\mathbf{w}^{+}\cdot\phi^{+}(t)
-\mathbf{w}^{-}\cdot\phi^{-}(t)),
\]
where $\phi^{+}$ includes progress, success, novelty, and dependency-support
signals, and $\phi^{-}$ includes failure, no-op, and redundancy signals. The
score is used only for pruning and diagnostics, not as verifier supervision.

Finally, contiguous retained steps with the same procedural intent are merged
into phase-level DAG nodes. Each node stores the inferred operation type,
task-relevant entities, compact supporting evidence, and optional skill
bindings. Retained dependencies are mapped into temporal, data-flow, or
control-flow edges; if no explicit dependency remains, the extractor falls back
to a temporal chain to preserve executability.

The extracted DAG is then used to update Skill-DAG Memory. Successful
executions insert new templates or reinforce matched ones, while successful
online replanning stores the revised trajectory as a new variant and lowers the
reliability of the original template. Failed executions do not create new
Skill-DAG templates and only update reliability statistics, preventing noisy or
unsuccessful behavior from being used as positive procedural memory.
\subsection{Cold-Start Deployment}

\model{} is designed to support both trajectory-rich and trajectory-scarce
deployment settings. The central question in cold-start deployment is whether
target-domain execution trajectories are available for training a
domain-specific Skill-GNN. If such trajectories are available, \model{}
constructs online-like graph training samples from trajectory prefixes and
trains an inductive Skill-GNN. If no target-domain trajectory is available,
\model{} performs a library-level warm start using only skill metadata to
construct pseudo task--skill pairs and weak skill--skill relations. During early
deployment, Atomic Skill Memory and available Skill-DAG templates provide the
primary guidance, while the warm-started Skill-GNN is used for bounded reranking.
Successful executions collected during deployment are then converted into
Skill-DAGs and transitions, which provide supervision for later Skill-GNN
adaptation.

A key reason why this transition is feasible is that Skill-GNN is inductive rather than transductive. It does not learn node embeddings tied to a fixed global skill graph. Instead, for each task state, \model{} dynamically constructs a candidate graph from retrieved atomic skills, task embeddings, state features, history features, and relation signals. Skill nodes are represented by semantic embeddings and execution-derived features, while message-passing parameters are shared across candidate graphs and edge types. Therefore, the trained Skill-GNN can be applied to unseen tasks, unseen candidate skill combinations, and newly extracted Skill-DAG variants as long as their features and relations can be constructed. This inductive design makes few-trajectory adaptation more practical in target domains.

In the trajectory-rich setting, each online-like sample is built from a prefix of a successful trajectory. The next skill is used as the target label, previous skills provide execution history, and a bounded candidate set is formed by combining the target skill, prefix-context skills, limited future positive skills, hard negatives retrieved from Atomic Skill Memory, and random negatives. The resulting graph contains task, state, history, and candidate skill features, but does not expose the full future DAG as model input. This prevents the model from relying on information unavailable during deployment.

In the trajectory-scarce setting, \model{} starts from a library-level warm
start rather than a trajectory-supervised graph model. Skill metadata are used
to build pseudo task--skill pairs and weak skill--skill relations, which provide
a lightweight initial structure for Skill-GNN. During early deployment, the
agent still relies primarily on Atomic Skill Memory retrieval and available
Skill-DAG templates, while the warm-started Skill-GNN is used only for bounded
reranking over retrieved candidates. As successful executions accumulate,
trajectory-DAG extraction converts them into reusable Skill-DAG templates and
training transitions, gradually adapting the system from retrieval-guided
cold start to structure-aware selection. Failed template executions are not
stored as new templates; they only reduce the reliability statistics of the
corresponding templates.

\subsection{Skill-GNN Training and Inference}
\begin{figure*}[t]
\centering
\includegraphics[width=\linewidth]{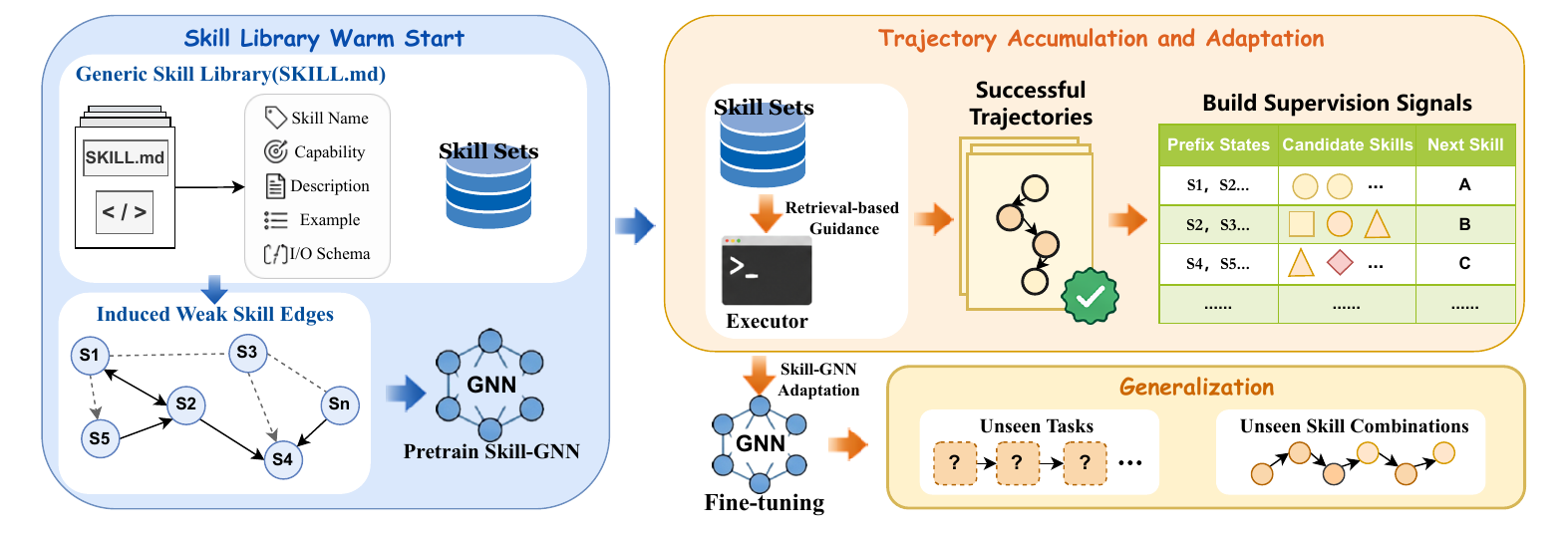}
\caption{Cold-start deployment of \model{}. When no target-domain trajectories
are available, \model{} first warm-starts Skill-GNN from the generic skill
library by constructing pseudo queries and weak skill relations from skill
metadata. As successful executions accumulate, trajectories are converted into
prefix-level supervision signals for further Skill-GNN adaptation, enabling
generalization to unseen tasks and unseen skill combinations.}
\label{fig:cold_start}
\end{figure*}
\label{app:skillgnn_training}

\paragraph{Candidate graph construction.}
Skill-GNN is applied to a task-conditioned candidate graph rather than to a
fixed global skill graph. Given a task query, the system first retrieves a
bounded candidate set from Atomic Skill Memory. The graph contains a task node,
optional state and history nodes, and one node for each candidate skill. Node
features concatenate the corresponding semantic embedding with lightweight
dynamic statistics, including empirical success rate, preference score, usage
count, and execution cost. The graph contains typed edges from the task/state
context to candidate skills, as well as skill-skill edges induced by semantic
relatedness or template dependencies when available. This construction makes
Skill-GNN inductive: new tasks and newly added skills can be handled by building
a fresh candidate graph at inference time.

\paragraph{Prediction heads.}
Skill-GNN encodes the candidate graph and produces both node-level and
graph-level representations. We use three lightweight heads on top of the GNN.
The next-skill head scores candidate skill nodes conditioned on the current
selected prefix, the value head predicts the utility of the candidate graph, and
the edge head predicts whether two skill nodes should be connected by a
dependency edge. During online inference, the next-skill scores are used to
rerank candidate atomic skills before injecting them into the agent context.

\paragraph{Trajectory-supervised training.}
When successful trajectories are available, we train Skill-GNN with
trajectory-prefix supervision. For a successful Skill-DAG with ordered nodes
$(v_1,\ldots,v_m)$, each prefix $(v_1,\ldots,v_{k-1})$ defines a training state,
and the next node $v_k$ is used as the target skill. The next-skill objective is
a cross-entropy loss over candidate skill nodes:
\[
\mathcal{L}_{\mathrm{next}}
=
-\log
\frac{\exp f_\theta(v_k \mid G, P_k)}
{\sum_{v \in \mathcal{C}}\exp f_\theta(v \mid G, P_k)},
\]
where $P_k$ is the selected prefix and $\mathcal{C}$ is the candidate skill set.
When available, we additionally use a value loss for graph-level success or
reward prediction and a binary edge loss for dependency prediction:
\[
\mathcal{L}
=
\lambda_{\mathrm{next}}\mathcal{L}_{\mathrm{next}}
+
\lambda_{\mathrm{value}}\mathcal{L}_{\mathrm{value}}
+
\lambda_{\mathrm{edge}}\mathcal{L}_{\mathrm{edge}}.
\]
In our implementation, $\mathcal{L}_{\mathrm{value}}$ is a smooth-$L_1$ loss and
$\mathcal{L}_{\mathrm{edge}}$ is a binary cross-entropy loss over candidate node
pairs.

\paragraph{Library-level cold-start pretraining.}
Figure~\ref{fig:cold_start} shows the cold-start path used when no
target-domain trajectory is available. Starting from a generic skill library,
\model{} constructs pseudo queries from each skill's name, one-line capability,
description, example tasks, tooling, and I/O schema. The corresponding skill is
used as the positive target, while semantically or lexically similar skills are
sampled as hard negatives. In parallel, weak skill-skill edges are induced from
library metadata, including semantic neighbors, shared artifacts, shared tools
or tags, and compatible input-output schemas. These pseudo queries and weak
edges provide initial next-skill and edge supervision for Skill-GNN before any
successful task trajectory has been collected.

\paragraph{Leakage prevention.}
The training sources are separated according to the deployment setting.
Library-level pretraining uses only skill metadata and does not use task
solutions, verifier rewards, or evaluation trajectories. Trajectory-supervised
training uses only trajectories from the training split or successful
post-deployment executions. In particular, diagnostic query--gold pairs used for
retrieval analysis are not used to train the model. This prevents evaluation
tasks from being converted into positive supervision before they are executed.
\begin{table*}[t]
\centering
\small
\begin{tabular}{lcccccc}
\toprule
Backbone & Val Acc@1 & Test Acc@1 & Top-3 & MRR & Edge-F1 & Inference ms \\
\midrule
GAT &
$0.9676{\pm}0.0042$ &
$0.9776{\pm}0.0018$ &
$0.9991{\pm}0.0010$ &
$0.9878{\pm}0.0007$ &
$0.9741{\pm}0.0027$ &
$4.57{\pm}0.23$ \\
GIN &
$\mathbf{0.9728{\pm}0.0007}$ &
$\mathbf{0.9808{\pm}0.0008}$ &
$0.9994{\pm}0.0004$ &
$\mathbf{0.9901{\pm}0.0006}$ &
$0.9771{\pm}0.0025$ &
$\mathbf{1.47{\pm}0.04}$ \\
Hetero-GNN &
$0.9683{\pm}0.0004$ &
$0.9783{\pm}0.0030$ &
$\mathbf{0.9995{\pm}0.0012}$ &
$0.9885{\pm}0.0017$ &
$\mathbf{0.9841{\pm}0.0009}$ &
$2.42{\pm}0.71$ \\
\bottomrule
\end{tabular}
\caption{Skill-GNN backbone comparison. Runtime is measured as milliseconds per inference sample.}
\label{tab:gnn-backbone}
\end{table*}

\begin{table*}[t]
\centering
\small
\begin{tabular}{llccccc}
\toprule
Study & Setting & Test Acc@1 & Top-3 & MRR & Edge-F1 & Inference ms \\
\midrule
Candidate Top-$K$ & $K=8$  & $0.9533{\pm}0.0016$ & $0.9698{\pm}0.0004$ & $0.9613{\pm}0.0008$ & $0.9553{\pm}0.0011$ & $3.82{\pm}2.06$ \\
Candidate Top-$K$ & $K=16$ & $0.9490{\pm}0.0029$ & $\mathbf{0.9960{\pm}0.0012}$ & $0.9588{\pm}0.0016$ & $0.9546{\pm}0.0009$ & $3.81{\pm}0.75$ \\
Candidate Top-$K$ & $K=24$ & $0.9467{\pm}0.0084$ & $0.9693{\pm}0.0007$ & $0.9576{\pm}0.0044$ & $0.9582{\pm}0.0018$ & $3.31{\pm}0.77$ \\
Candidate Top-$K$ & $K=32$ & $\mathbf{0.9546{\pm}0.0021}$ & $\mathbf{0.9970{\pm}0.0008}$ & $\mathbf{0.9621{\pm}0.0011}$ & $\mathbf{0.9591{\pm}0.0014}$ & $\mathbf{2.58{\pm}0.21}$ \\
\midrule
Layers & 1 & $\mathbf{0.9520{\pm}0.0014}$ & $0.9695{\pm}0.0008$ & $\mathbf{0.9605{\pm}0.0012}$ & $0.9537{\pm}0.0029$ & $\mathbf{1.74{\pm}0.04}$ \\
Layers & 2 & $0.9490{\pm}0.0029$ & $\mathbf{0.9955{\pm}0.0010}$ & $0.9588{\pm}0.0016$ & $\mathbf{0.9546{\pm}0.0009}$ & $2.13{\pm}0.09$ \\
Layers & 3 & $0.9518{\pm}0.0026$ & $0.9695{\pm}0.0004$ & $0.9602{\pm}0.0016$ & $0.9503{\pm}0.0011$ & $2.73{\pm}0.06$ \\
\midrule
Hidden Dim & 64  & $\mathbf{0.9493{\pm}0.0044}$ & $0.9693{\pm}0.0007$ & $\mathbf{0.9589{\pm}0.0022}$ & $0.9519{\pm}0.0010$ & $\mathbf{2.12{\pm}0.08}$ \\
Hidden Dim & 128 & $0.9490{\pm}0.0029$ & $\mathbf{0.9965{\pm}0.0009}$ & $0.9588{\pm}0.0016$ & $\mathbf{0.9546{\pm}0.0009}$ & $2.16{\pm}0.05$ \\
Hidden Dim & 256 & $0.9483{\pm}0.0056$ & $0.9698{\pm}0.0004$ & $0.9588{\pm}0.0028$ & $0.9497{\pm}0.0052$ & $2.53{\pm}0.11$ \\
\midrule
Edge Neg. Ratio & 1 & $\mathbf{0.9520{\pm}0.0016}$ & $0.9698{\pm}0.0004$ & $\mathbf{0.9605{\pm}0.0007}$ & $\mathbf{0.9625{\pm}0.0009}$ & $2.62{\pm}0.32$ \\
Edge Neg. Ratio & 3 & $0.9490{\pm}0.0029$ & $\mathbf{0.9960{\pm}0.0010}$ & $0.9588{\pm}0.0016$ & $0.9546{\pm}0.0009$ & $\mathbf{2.45{\pm}0.19}$ \\
Edge Neg. Ratio & 5 & $0.9514{\pm}0.0010$ & $\mathbf{0.9970{\pm}0.0008}$ & $0.9601{\pm}0.0005$ & $0.9457{\pm}0.0046$ & $3.20{\pm}0.72$ \\
\bottomrule
\end{tabular}
\caption{Skill-GNN hyperparameter sensitivity. Results show that performance remains stable across candidate size, depth, hidden dimension, and edge negative sampling ratio.}
\label{tab:gnn-hyperparams}
\end{table*}

\begin{figure*}[t]
\centering
\includegraphics[width=\textwidth]{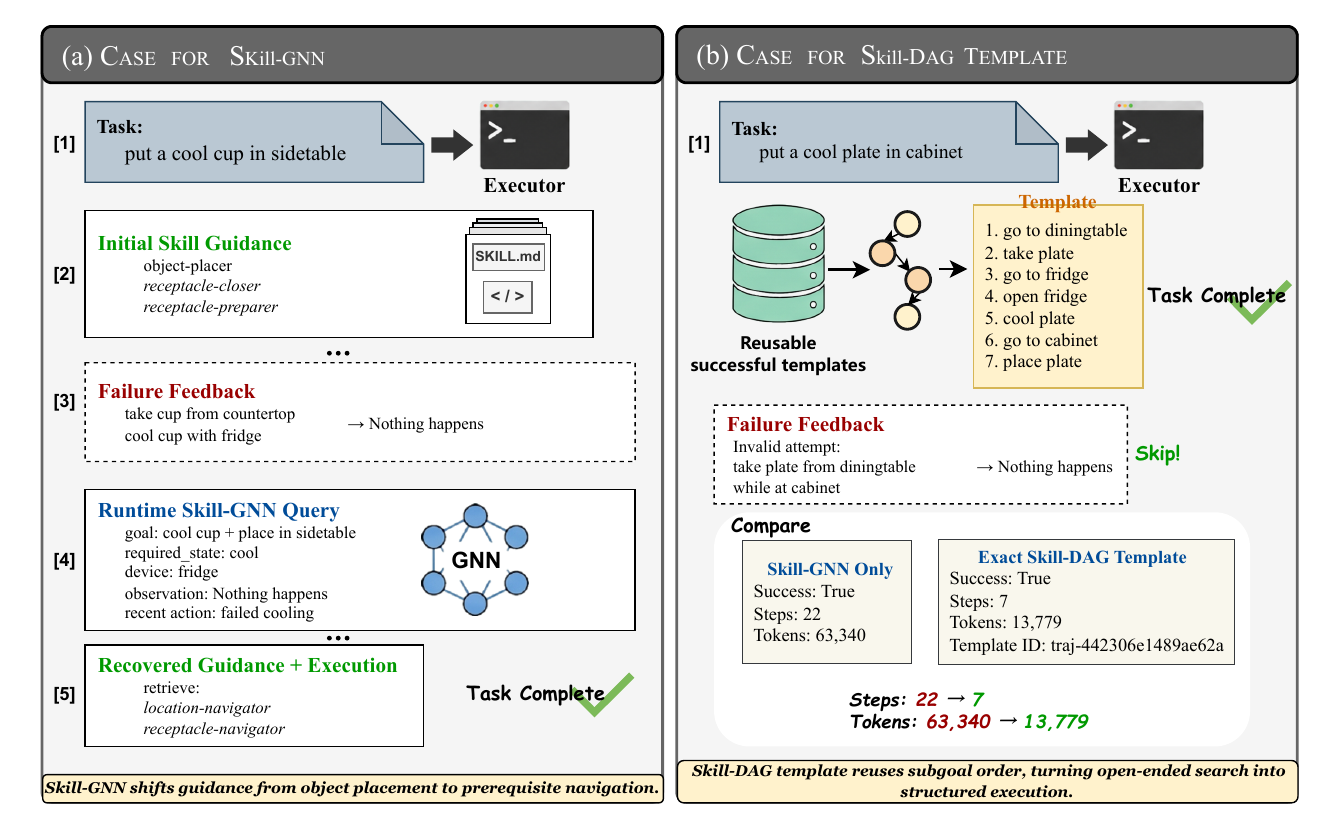}
\caption{ALFWorld case studies illustrating the complementary roles of
Skill-GNN and Skill-DAG templates. As shown in (a), Skill-GNN provides
state-conditioned recovery guidance after a failed action by shifting the
agent's attention from object placement to prerequisite navigation. As shown in
(b), a retrieved Skill-DAG template reuses a compact successful subgoal order,
reducing redundant exploration and lowering both steps and token usage.}
\label{fig:case_studies}
\end{figure*}

\section{Experimental Details}
\label{app:experimental_details}

\paragraph{Fairness protocol.}
Across methods, we keep the benchmark split, skill library, agent interface,
execution environment, verifier, and evaluation script fixed whenever
applicable. Differences between methods are restricted to how skills are
selected, structured, and injected into the agent context. Offline memory
construction and Skill-GNN training are performed before evaluation and are not
counted as online execution time.

\paragraph{ALFWorld configuration.}
For ALFWorld, all skill-based methods use the same 37-skill library.
Atomic Skill Memory is constructed directly from the corresponding skill
documents. Skill-DAG Memory is constructed only from successful trajectories
outside the evaluation episodes. Skill-GNN is trained with trajectory-prefix
supervision: for each successful trajectory, a prefix of executed skills defines
the current procedural state, and the next executed skill is used as the
prediction target. The GNN is evaluated on disjoint development episodes and
does not access evaluation trajectories during memory construction or training.
The maximum number of environment interaction steps is set to 30.

\paragraph{SkillsBench configuration.}
For SkillsBench, all compared skill-based methods use the same
\texttt{skills\_200} library and the same Harbor-based agent execution
interface. Each task is allowed at most two attempts, with two concurrent trials,
following the GoS execution setting. Atomic Skill Memory is constructed from
skill names, capability summaries, descriptions, metadata, examples, and
interface schemas. Because target-domain successful trajectories may be
unavailable before evaluation, \model{} uses the library-level cold-start
procedure described above. Specifically, pseudo task--skill pairs are
constructed from skill metadata, and weak skill--skill edges are induced from
semantic similarity, shared tools or artifacts, tags, and schema compatibility.
When successful executions are later accumulated, their extracted Skill-DAGs can
be used for further Skill-GNN adaptation.

\paragraph{Key implementation parameters.}
Table~\ref{tab:key_impl_params} reports the main Skill-GNN implementation
parameters used in the trajectory-supervised setting. In ALFWorld, Skill-GNN is
trained from successful non-evaluation trajectories with trajectory-prefix
supervision. In SkillsBench, target-domain trajectories are scarce before
evaluation, so \model{} uses a library-level warm start constructed only from
skill metadata, pseudo task--skill pairs, and weak skill--skill relations. When
a small number of successful deployment trajectories become available, they can
be used for further adaptation, but evaluation-task trajectories are never used
for training. All runtime parameters are fixed within each benchmark and are not
tuned for individual tasks.

\begin{table}[t]
\centering
\small
\setlength{\tabcolsep}{5pt}
\renewcommand{\arraystretch}{1.08}
\begin{tabular}{ll}
\toprule
Parameter & Value \\
\midrule
GNN backbone & Hetero-GNN \\
Hidden dimension & 128 \\
GNN layers & 2 \\
Optimizer & AdamW \\
Learning rate & $1\times10^{-3}$ \\
Weight decay & $1\times10^{-5}$ \\
Batch size & 64 \\
Epochs & 5 \\
Template threshold & 0.80 \\
Decoding temperature & Benchmark default \\
\bottomrule
\end{tabular}
\caption{Key Skill-GNN implementation parameters in the trajectory-supervised
setting.}
\label{tab:key_impl_params}
\end{table}

\paragraph{Token and runtime accounting.}
For ALFWorld, token usage and runtime are averaged over evaluated episodes.
For SkillsBench, $T$ denotes raw input tokens rather than discounted
cached-token cost. $S$ measures only the agent execution phase. It includes
online skill retrieval, Skill-GNN reranking, and prompt construction when they
are invoked during agent execution, but excludes Docker image build time,
environment setup, dependency installation, and verifier runtime.

\paragraph{Leakage prevention.}
Library-level warm start uses only skill metadata and does not use task
solutions, verifier rewards, or evaluation trajectories. Trajectory-supervised
Skill-GNN training uses only training trajectories or successful
post-deployment executions. Diagnostic query--gold pairs used for retrieval
analysis are used only for analysis and are not used to train the model. Failed
executions do not create positive Skill-DAG templates and only update reliability
statistics.

\section{Additional Results}

\label{app:additional_results}

\subsection{Skill-GNN Backbone Comparison}
\label{app:gnn_backbone}

Table~\ref{tab:gnn-backbone} compares different GNN backbones in an offline
next-skill prediction setting. This diagnostic evaluates whether the model can
rank the correct next skill within a bounded candidate graph, and should not be
interpreted as end-to-end task success. All backbones achieve high accuracy,
suggesting that the constructed candidate graphs contain strong task-skill
signals. GIN obtains the fastest inference speed and slightly higher Acc@1,
while Hetero-GNN achieves the best Edge-F1 and perfect Top-3 accuracy. We use
Hetero-GNN in the main experiments because its typed message passing better
matches the heterogeneous task--skill graph and provides stronger dependency
modeling, which is important when Skill-DAG templates or typed skill relations
are available.

\subsection{Hyperparameter Sensitivity}
\label{app:gnn_hyperparams}

Table~\ref{tab:gnn-hyperparams} reports sensitivity analyses for candidate set
size, GNN depth, hidden dimension, and edge negative sampling ratio. The results
are relatively stable across settings, indicating that Skill-GNN does not rely
on a narrowly tuned hyperparameter configuration. Increasing the candidate set
to $K=32$ improves Acc@1, MRR, and Edge-F1 while keeping inference efficient,
suggesting that a moderately larger candidate graph provides useful distractors
without overwhelming the reranker. Increasing depth or hidden size does not
consistently improve performance, which implies that the candidate graphs are
compact and that most useful information can be captured with a lightweight
GNN. For edge supervision, a small negative ratio performs best on Edge-F1,
while larger ratios slightly reduce edge prediction quality. We therefore use a
moderate configuration in the main experiments to balance accuracy, dependency
modeling, and inference overhead.

These offline results mainly validate the ranking and structural modeling
capacity of Skill-GNN. They do not replace end-to-end benchmark evaluation,
where performance also depends on LLM reasoning, action execution, environment
feedback, and verifier-specific requirements.

\section{Case Studies}
\label{app:case_studies}
Figure~\ref{fig:case_studies} presents two ALFWorld case studies that illustrate
the complementary roles of Skill-GNN and Skill-DAG templates. The first case
shows how Skill-GNN supports state-conditioned recovery during execution, while
the second case shows how a Skill-DAG template provides reusable procedural
structure before execution.

\paragraph{Skill-GNN for state-conditioned recovery.}
As shown in Figure~\ref{fig:case_studies}(a), in the task
\textit{put a cool cup in sidetable}, the agent initially receives skill
guidance related to object placement and receptacle preparation. This guidance
captures the general structure of the task, but does not fully resolve the
cooling precondition. After taking the cup, the agent attempts to execute
\texttt{cool cup with fridge} before navigating to the refrigerator, and the
environment returns \texttt{Nothing happens}. The failure feedback triggers a
runtime Skill-GNN query that includes the task goal, required state, target
device, current observation, and recent failed action. Skill-GNN then retrieves
navigation-oriented skills, such as location navigation and receptacle
navigation. With this updated guidance, the agent moves to the refrigerator,
opens it, cools the cup, navigates to the sidetable, and completes the task.
This case shows that Skill-GNN is not merely an initial retriever; it can also
revise skill guidance according to the current execution state and failure
feedback.

\paragraph{Skill-DAG templates for procedural reuse.}
As shown in Figure~\ref{fig:case_studies}(b), in the task
\textit{put a cool plate in cabinet}, Skill-GNN alone eventually succeeds but
requires 22 steps and 63,340 tokens. The trajectory includes redundant search
and invalid action attempts caused by unresolved preconditions or mismatched
locations. In contrast, the retrieved Skill-DAG template provides a compact
successful procedure: go to the dining table, take the plate, go to the
refrigerator, open it, cool the plate, go to the cabinet, and place the plate.
This template-guided execution succeeds in 7 steps with 13,779 tokens. The
example demonstrates that Skill-DAG templates provide more than relevant skill
selection: they encode a reusable ordering of subgoals, turning open-ended
exploration into structured execution.

\paragraph{Discussion.}
These two cases highlight different forms of structure-aware skill use.
Skill-GNN answers which skill is relevant under the current state, especially
after failure feedback changes the agent's immediate needs. Skill-DAG templates
answer how subgoals should be ordered by reusing successful procedural
skeletons. Together, the two mechanisms support both adaptive recovery and
efficient long-horizon execution.

\bibliographystyle{aaai2027}